\documentclass[10pt,twocolumn,letterpaper]{article}

\usepackage{iccv}
\usepackage{times}
\usepackage{epsfig}
\usepackage{graphicx}
\usepackage{amsmath}
\usepackage{amssymb}
\usepackage{color}

\usepackage[pagebackref=true,breaklinks=true,letterpaper=true,colorlinks,bookmarks=false]{hyperref}

\iccvfinalcopy 

\def\iccvPaperID{1637} 

\ificcvfinal\fi
\begin{document}

\title{Automatic Concept Discovery from Parallel Text and Visual Corpora}

\author{Chen Sun\\
Univ. of Southern California\\
{\tt\small chensun@usc.edu}
\and
Chuang Gan\thanks{This work was done when Chuang Gan was a visiting researcher at University of Southern California.}\\
Tsinghua University\\
{\tt\small ganchuang1990@gmail.com}
\and
Ram Nevatia\\
Univ. of Southern California\\
{\tt\small nevatia@usc.edu}
}

\maketitle

\begin{abstract}
 Humans connect language and vision to perceive the world. How to build a similar connection for computers? One possible way is via visual concepts, which are text terms that relate to visually discriminative entities. We propose an automatic visual concept discovery algorithm using parallel text and visual corpora; it filters text terms based on the visual discriminative power of the associated images, and groups them into concepts using visual and semantic similarities. We illustrate the applications of the discovered concepts using bidirectional image and sentence retrieval task and image tagging task, and show that the discovered concepts not only outperform several large sets of manually selected concepts significantly, but also achieves the state-of-the-art performance in the retrieval task.
\end{abstract}

\section{Introduction}




Language and vision are both important for us to understand the world. Humans are good at connecting the two modalities. Consider the sentence ``A fluffy dog leaps to catch a ball'': we can all relate \textit{fluffy dog}, \textit{dog leap} and \textit{catch ball} to the visual world and describe them in our own words easily. However, to enable a computer to do something similar, we need to first understand what to learn from the visual world, and how to relate them to the text world.

\textit{Visual concepts} are a natural choice to serve as the basic unit to connect language and vision. A visual concept is a subset of human vocabulary which specifies a group of visual entities (e.g. \textit{fluffy dog, curly dog}). We name the collection of visual concepts as a visual vocabulary. Computer vision researchers have long collected image examples of manually selected visual concepts, and used them to train concept detectors. For example, ImageNet~\cite{imagenet_cvpr09} selects 21,841 synsets in WordNet as the visual concepts, and has by far collected 14,197,122 images in total. One limitation of the manually selected concepts is that their visual detectors often fail to capture the complexity of the visual world, and cannot adapt to different domains. For example, people may be interested in detecting \textit{birthday cakes} when they try to identify a \textit{birthday party}, but this concept is not present in ImageNet. 


To address this problem, we propose to discover the visual concepts automatically by joint use of parallel text and visual corpora. The text data in parallel corpora offers a rich set of terms humans use to describe visual entities, while visual data has the potential to help computer organize the terms into visual concepts. To be useful, we argue that the visual concepts should have the following properties:

\textbf{Discriminative}: a visual concept must refer to visually discriminative entities that can be learned by available computer vision algorithms.

\textbf{Compact}: different terms describing the same set of visual entities should be merged into a single concept.

\begin{figure*}[t]
\begin{center}
\begin{tabular}{c}
\includegraphics[width=0.98\linewidth]{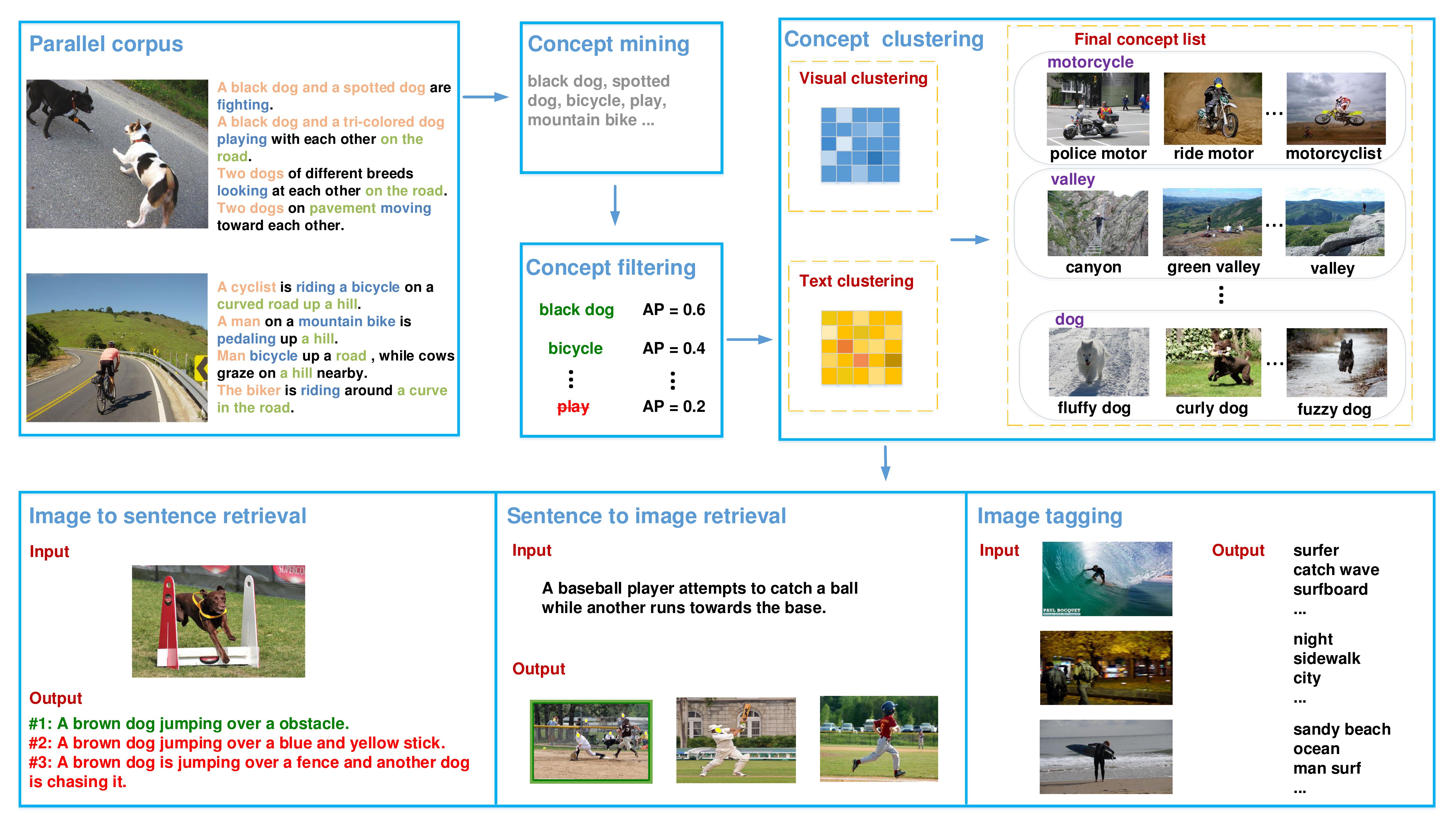}\\
\end{tabular}
\end{center}
\caption{Overview of the concept discovery framework. Given a parallel corpus of images and their descriptions, we first extract unigrams and dependency bigrams from the text data. These terms are filtered with the cross validation average precision (AP) trained on their associated images. The remaining terms are grouped into concept clusters based on both visual and semantic similarity.}
\label{fig:teaser}
\end{figure*}


Our proposed visual concept discovery (VCD) framework first extracts unigrams and dependencies from the text data. It then computes the visual discriminative power of these terms using their associated images and filters out the terms with low cross-validated average precision. The remaining terms may be merged together if they correspond to very similar visual entities. To achieve this, we use semantic similarity and visual similarity scores, and cluster terms based on these similarities. The final output of VCD is a concept vocabulary, where each concept consists a set of terms and has a set of associated images. The pipeline of our approach is illustrated in Figure~\ref{fig:teaser}.

We work with the Flickr 8k data set to discover visual concepts; it consists of 8,000 images downloaded from the Flickr website. Each image was annotated by 5 Amazon Mechanical Turk (AMT) workers to describe its content. We design a concept based pipeline for bidirectional image and sentence retrieval task~\cite{DBLP:journals/jair/HodoshYH13} to automatically evaluate the quality of the discovered concepts. We also conduct a human evaluation on a free-form image tagging task using visual concepts. Evaluation results show that the discovered concepts outperform manually selected concepts significantly.


Our key contributions include: 
\begin{itemize}
\item We show that manually selected concepts often fail to capture the complexity of and to evolve with the visual world; 
\item We propose the VCD framework, which automatically generates discriminative and compact visual vocabularies from parallel corpora; 
\item We demonstrate qualitatively and quantitatively that the discovered concepts outperform several large sets of manually selected concepts significantly. They also perform competitively in the image sentence retrieval task against state-of-the-art embedding based approaches.
\end{itemize}

\section{Related Work}
\label{sec:related}

\textbf{Applications of visual concepts.}
Visual concepts have been widely used in visual recognition tasks~\cite{LiSXL10OB,Sadanand12AB,zhou2014places}. For example, \cite{DBLP:conf/cvpr/FarhadiEHF09} addresses the problem of describing objects with pre-defined attributes. Sadeghi et al.~\cite{DBLP:conf/cvpr/SadeghiF11} propose to recognize complex visual composites by defining visual phrases. For video analysis, people commonly use predefined pools of concepts (e.g. \textit{blowing candle}, \textit{cutting cake}) to help classify and describe high-level activities or events (e.g. \textit{birthday party})~\cite{sun_nevatia_cvpr14}. However, their concept vocabularies are usually manually selected.

\textbf{Concept naming and accuracy-specificity trade-off.} Visual concepts can be categorized~\cite{rosch1978} and organized as a hierarchy where the leaves are the most specific and the root is the most general. For example, ImageNet concepts~\cite{imagenet_cvpr09} are organized following the rule-based WordNet~\cite{Miller95wordnet:a} hierarchy. Similar structure also exists for actions~\cite{caba2015activitynet}. Since concept classification is not always reliable, Deng et al.~\cite{DengKrauseBergFei-Fei_CVPR2012} propose a method to allow accuracy-specificity trade-off of object concepts on WordNet. As WordNet synsets do not always correspond to how people name the concepts, Ordonez et al.~\cite{DBLP:conf/iccv/OrdonezDCBB13} study the problem of entry-level category prediction by collecting \textit{natural categories} from humans.


\textbf{Concept learning from web data.}
Our research is closely related to the recent work on visual data collection from web images~\cite{wu2015webconcept,chen2013neil,DBLP:conf/cvpr/DivvalaFG14,DBLP:conf/eccv/GolgeD14} or weakly annotated videos~\cite{chen2013watching}. Their goal is to collect training images from the Internet with minimum human supervision, but for pre-defined concepts. In particular, NEIL~\cite{chen2013neil} starts with a few exemplar images per concept, and iteratively refines its concept detectors using image search results. LEVAN~\cite{DBLP:conf/cvpr/DivvalaFG14} explores the sub-categories of a given concept by mining bigrams from large text corpus and using the bigrams to retrieve training images from image search engines. Recently, Zhou et al.~\cite{zhou2015conceptlearner} use noisily tagged Flickr images to train concept detectors, but do not consider the semantic similarity among different tags. Our VCD framework is able to generate the concept vocabulary for them to learn detectors.

\textbf{Sentence generation and retrieval for images.} Image descriptions can be generated by detection or retrieval. The detection based approach usually defines a set of visual concepts (e.g. \textit{objects, actions} and \textit{scenes}), learns concept detectors and use the top detected concepts to generate sentences. The sentences can be generated using templates~\cite{guadarrama:ICCV13,SunECCV,Kulkarni11babytalk} or language models~\cite{rohrbachtranslating,DBLP:conf/acl/KuznetsovaOBBC13}. The performance of detection is often limited by missing concepts and inaccurate concept detectors. Retrieval-based sentence generation~\cite{ordonez2011im2text,kuznetsova2012collective,DBLP:journals/tacl/YoungLHH14} works by retrieving sentences or sentence components from an existing pool of sentence and image pairs, and use them for description. The retrieval criteria is usually based on the visual similarity of image features. To allow bidirectional retrieval of sentences and images, several work~\cite{DBLP:journals/jair/HodoshYH13,DBLP:conf/eccv/GongWHHL14,devise} embed image and text raw features into a common latent space using methods like Kernel Canonical Component Analysis~\cite{DBLP:dblp_journals/jmlr/BachJ02}. There is also a trend to embed sentences with recurrent neural networks (RNN)~\cite{DBLP:journals/tacl/SocherKLMN14,DBLP:journals/corr/KarpathyF14,DBLP:journals/corr/MaoXYWY14,DBLP:journals/corr/ChenZ14a,DBLP:journals/corr/KirosSZ14}, which achieves the state-of-the-art performance in sentence retrieval and generation tasks.

\section{Visual Concept Discovery Pipeline}
\label{sec:pipeline}
This section describes the VCD pipeline. Given
a parallel corpus with images and their text descriptions, we
first mine the text data to select candidate concepts. Due to the
diversity of both visual world and human language, the pool of
candidate concepts is large. We use visual data to filter the terms
which are not visually discriminative, and then group the remaining terms into compact concept clusters.

\subsection{Concept Mining From Sentences}

To collect the candidate concepts, we use unigrams as well as the grammatical relations called \textit{dependencies}~\cite{Marneffe06generatingtyped}. Unlike the syntax tree based representation of sentences, dependencies operate directly on pairs of words. Consider a simple sentence \textit{``a little boy is riding a white horse''}, \textit{white horse} and \textit{little boy} belong to the adjective modifier (\textit{amod}) dependency, and \textit{ride horse} belongs to the direct object (\textit{dobj}) dependency. As the number of dependency types is large, we manually select a subset of 9 types which are likely to correspond to visual concepts. The selected dependency types are: \textit{acomp, agent, amod, dobj, iobj, nsubj, nsubjpass, prt} and
\textit{vmod}.

The concept mining process proceeds as follows: we first parse the
sentences in the parallel corpus with the Stanford CoreNLP
parser~\cite{Marneffe06generatingtyped}, and collect the terms with
the interesting dependency types. We also select unigrams which are
annotated as \textit{noun, verb, adjective} and \textit{adverb} by a
part-of-speech tagger. We use the \textit{lemmatized} form of the
selected unigrams and phrases such that nouns in singular and plural
forms and verbs in different tenses are grouped together. After
parsing the whole corpus, we remove the terms which
occur fewer than $k$ times.

\begin{table}
\small
  \begin{center}
    \begin{tabular}{c|c}
      \hline
      Preserved terms & Filtered terms\\
      \hline
      play tennis, play basketball & play\\
      \hline
      bench, kayak & red bench, blue kayak\\
      \hline
      sheer, tri-colored & real, Mexican\\
      \hline
      biker, dog & cigar, chess\\
      \hline
     \end{tabular}
  \end{center}
  \caption{Preserved and filtered terms from Flickr 8k data set. A term might be filtered if it's \textit{abstract} (first row), \textit{too detailed} (second row) or not \textit{visually discriminative} (third row). Sometimes our algorithm may filter out visual entities which are difficult to recognize (final row).}
  \label{tab:filter_example}
\end{table}

\subsection{Concept Filtering and Clustering}
\label{sec:cluster}
The unigrams and dependencies selected from text data contain terms
which may not have concrete visual patterns or may not be easy to
learn with visual features. The images in the parallel
corpora are helpful to filter out these terms. We represent images using feature activations from pre-trained deep convolutional neural networks (CNN), they are image-level holistic features.

Since the number of terms mined from text data is
large, the concept filtering algorithm needs to be efficient. For the
images associated with a certain term, we do a
2-fold cross validation with a linear SVM, using randomly sampled
negative training data. We compute average precision (AP) on
cross-validated results, and remove the terms with AP lower than a
threshold. Some of the preserved and filtered terms are listed in Table~\ref{tab:filter_example}.

Many of the remaining terms are synonyms (e.g. \textit{ride bicycle} and \textit{ride bike}). These terms are likely to confuse the concept classifier training algorithm. It is important to merge them together to make the concept set more compact. Besides, although some terms refer to different visual entities, they are similar visually and semantically (e.g. a \textit{red jersey} and a \textit{orange jersey}); it is often beneficial to group them together to have more image examples for training. This motivates us to cluster the concepts based on both visual similarity and semantic similarity.

\textbf{Visual similarity:} We use the holistic image
features to measure visual similarity between different candidate
concept terms. We learn two classifiers $f_{t_1}$ and $f_{t_2}$ for terms $t_1$ and $t_2$ using their associated image sets $I_{t_1}$ and $I_{t_2}$;  negative data is randomly sampled from those not associated with $t_1$ and $t_2$. To measure the similarity from $t_1$ to $t_2$, we compute the median of classifier $f_{t_1}$'s response on the positive samples of $t_2$.

\begin{equation}
\widehat{S}_v(t_1, t_2) = \textrm{median}_{I \in I_{t_2}}(f_{t_1}(I))
\end{equation}
\begin{equation}
S_v(t_1, t_2) = \textrm{min}\left(\widehat{S}_v(t_1, t_2), \widehat{S}_v(t_2, t_1)\right)
\end{equation}
Here the outputs of $f_{t_1}$ are normalized to $[0,1]$ by a
Sigmoid function. We take the minimum of $\widehat{S}_v(t_1, t_2)$ and
$\widehat{S}_v(t_2, t_1)$ to make it a symmetric similarity measurement.

The intuition behind this similarity measurement is that visual instances
associated with a term are more likely to get high scores from the
classifiers of other visually similar terms.

\textbf{Semantic similarity:} We also measure the similarity of two
terms in the semantic space, which are computed by data-driven word embeddings. In particular, we train a skip-gram model~\cite{word2vec} using the
English dump of Wikipedia. The basic idea of skip-gram model is to fit the word embeddings such that the words in corpus can predict their context with high probability. Semantically similar words lie close to each other in the embedded space.


Word embedding algorithm assigns a $D$-dimension vector for each word
in the vocabulary. For dependencies, we take the average of the word vectors
from each word of the dependency, and $L2$-normalize the averaged vector. The semantic similarity $S_w(t_1,t_2)$ of two candidate concept terms $t_1$ and $t_2$ is defined as the cosine similarity of their word embeddings.

\textbf{Concept clustering:} Denote the visual similarity matrix as
$\mathcal{S}_v$ and the semantic similarity matrix as $\mathcal{S}_w$,
we compute the overall similarity matrix by
\begin{equation}
\mathcal{S} = \mathcal{S}_v^\lambda \cdot \mathcal{S}_w^{1-\lambda}
\end{equation}
where $\cdot$ is element-wise matrix multiplication and $\lambda \in [0,1]$ is a parameter controlling the weight assigned to visual similarity.

We then use spectral clustering to cluster the candidate
concept terms into $K$ concept groups. It is a natural choice when similarity matrix is available. We use the algorithm implemented in
the Python SKLearn toolkit, fix the eigen solver to \textit{arpack}
and assign the labels with K-means. 

After the clustering stage, each concept is represented as a set of
terms, as well as their associated visual instances. One can use the associated visual instances to train concept detectors with SVM or neural networks.

\begin{table}
  \begin{center}
  \small
    \begin{tabular}{c|c}
      \hline
      Type & Concept terms\\
      \hline
      Object & \{jersey, red jersey, orange jersey\}\\
      \hline      
      Activity & \{dribble, player dribble, dribble ball\}\\
      \hline
      Attribute & \{mountainous, hilly\}\\
      \hline
      Scene & \{blue water, clear water, green water\}\\
      \hline
      Mixed & \{swimming, diving, pool, blue pool\}\\
      \hline
      Mixed & \{ride bull, rodeo, buck, bull\}\\
      \hline
     \end{tabular}
  \end{center}
  \caption{Concepts discovered by our framework from Flickr 8k data set.}
  \label{tab:cluster_example}
\end{table}

\subsection{Discussion}

Table~\ref{tab:cluster_example} shows some of the concepts discovered by our framework. It can automatically generate concepts related to objects, attributes, scenes and activities, and identify the different terms associated with each concept. We observe that sometimes a more general term (\textit{jersey}) is merged with a more specific term (\textit{red jersey}) due to high visual similarity. 

We also find that there are some mixed type concepts of objects, activities and scenes. For example, \textit{swimming} and \textit{pool} belongs to the same concept, possibly due to their high co-occurrence rate. One extreme case is that \textit{German} and \textit{German Shepherd} are grouped together as the two words always occur together in the training data. We believe the problem can be mitigated by using a larger parallel corpus.

\begin{table}
  \begin{center}
  \small
    \begin{tabular}{c|c}
      \hline
      $\lambda$ & Concept terms\\
      \hline
      0 & \{wedding, church\}, \{skyscraper, tall building\}\\
      \hline      
      1 & \{skyscraper, church\}, \{wedding, birthday\}\\
      \hline
      0.3 & \{wedding, bridal party\}, \{church\}, \{skyscraper\}\\
      \hline
     \end{tabular}
  \end{center}
  \caption{Different $\lambda$ affects the term groupings in the discovered concepts. Total concept number is fixed to 1,200.}
  \label{tab:cluster_lambda}
\end{table}

Table~\ref{tab:cluster_lambda} shows different concept clusters when semantic similarity is ignored ($\lambda=0$), dominant ($\lambda=1$) and combined with visual similarity. As expected, when $\lambda$ is small, terms that look similar or often co-occur in images tend to be grouped together. As our semantic similarity is based on word co-occurrence, ignoring visual similarity may lead to sub-optimal concept clusters such as \textit{wedding} and \textit{birthday}.

\section{Concept Based Image and Sentence Retrieval}
Consider a set of images, each of which has a few ground truth sentence annotations, the goal of bidirectional retrieval is to learn a ranking function from image to sentence and vice versa, such that the ground truth entries rank at the top of the retrieved list. Many previous methods approach the task by learning embeddings from raw feature space~\cite{DBLP:journals/jair/HodoshYH13,DBLP:journals/corr/KarpathyJF14,DBLP:conf/eccv/GongWHHL14}.

We propose an alternative approach to the embedding based methods which uses concept space directly. Let's start with the sentence to image direction. With the discovered concepts, this problem can be approached by two steps: first, identify the concepts from the sentences; second, select the images with highest responses for those concepts. Suppose we take the sum of the concept responses, this is equivalent to projecting the sentence into the same concept-based space as images, and measuring the image sentence similarity by an inner product. This formulation allows us to use the same similarity function for image to sentence and sentence to image retrieval.

\textbf{Sentence mapping:} Mapping a sentence to the concept space is
straightforward. We run the same parser as used in concept mining to collect terms. Remember that each concept is represented as a set of terms: denote the term set for the incoming sentence as $\mathcal{T}=\{t_1,t_2,...,t_N\}$, and the term set for concept $i$ as $\mathcal{C}_i = \{c^i_1,c^i_2,...,c^i_M\}$, we have the sentence's response for $\mathcal{C}_i$ as
\begin{equation}
\phi_i(\mathcal{T}) = \max_{t\in \mathcal{T}, c\in \mathcal{C}_i}\delta(t,c)
\end{equation}
Here $\delta(t,c)$ is a function that measures the similarity between $t$
and $c$. We set $\delta(t,c)=1$ if the cosine similarity of $t$ and $c$'s word embedding is greater than a certain threshold, and $0$ otherwise. In practice we set the threshold to $0.95$.

There are some common concepts which occur in most of the sentences
(e.g. a \textit{person}); to down-weight these common concepts, we
normalize the scores with term frequency-inverse
document frequency (tf-idf), learned from the training text corpus.

\textbf{Image mapping:} To measure the response of an image to a certain concept, we need to collect its positive and negative examples. For concepts discovered from parallel corpora, we have their associated images. The set of training images can be augmented with existing image data sets or by manual annotation.

Assume that training images are ready and concept classifiers have been trained, we then compute the continuous classifier scores for
an image over all concepts, and normalize each of them to be
$[-1,1]$. The normalization step is important as using
non-negative confidence scores biases the system towards longer sentences.

Since image and text data are mapped into a common concept space, the performance of bidirectional retrieval depends on: (1) whether the concept vocabulary covers the terms and visual entities used in query data; (2) whether concept detectors are powerful enough to extract useful information from visual data. It is thus useful to evaluate the quality of discovered concepts against existing concept vocabularies and their concept detectors.


\section{Evaluation}
\label{sec:exp}
In this section, we first evaluate our proposed concept discovery pipeline based on the bidirectional sentence image retrieval task. We use the discovered concepts to generate concept-based image descriptions, and report human evaluation results.

\subsection{Bidirectional Sentence Image Retrieval}
\label{sec:flickr}

\textbf{Data:} We use 6,000 images from the Flickr 8k~\cite{DBLP:journals/jair/HodoshYH13} data set for training, 1,000 images for validation and another 1,000 for testing. We use all 5 sentences per image for both training and testing. Flickr 30k~\cite{DBLP:journals/tacl/YoungLHH14} is an extended version of Flickr 8k. We select 29,000 images (no overlap to the testing images) to study whether more training data yields better concept detectors. We also report results when the visual concept discovery, concept detector training and evaluation are all conducted on Flickr 30k. For this purpose, we use the standard setting~\cite{DBLP:journals/corr/KarpathyF14,DBLP:journals/corr/KirosSZ14} where 29,000 images are used for training, 1,000 images for validation and 1,000 images for testing. Again, each image comes with 5 sentences. Finally, we randomly select 1,000 images from the lately released Microsoft COCO~\cite{DBLP:conf/eccv/LinMBHPRDZ14} data set to study if the discovered concept vocabulary and associated classifiers generalize to another data set. 

\textbf{Evaluation metric:} Recall$@k$ is used for evaluation. It computes the percentage of ground truth entries ranked in the top $k$ retrieved results, over all queries. We also report median rank of the first retrieved ground truth entries.

\textbf{Image representation and classifier training:} Similar to \cite{DBLP:conf/eccv/GongWHHL14,DBLP:journals/corr/KirosSZ14}, we extracted CNN activations as image-level features; such features have shown state-of-the-art performance in recent object recognition results~\cite{AlexNet,girshick14CVPR}. We adapted the CNN implementation provided by Caffe~\cite{jia2014caffe}, and used the 19-layer network architecture and parameters from Oxford~\cite{DBLP:journals/corr/SimonyanZ14a}. The feature activations from the network's first fully-connected layer fc6 were used as image representations, each of which has 4,096 dimensions.

To train concept classifiers, we normalized the feature activations with $L2$-norm. We randomly sampled 1,000 images as negative data. We used the linear SVM~\cite{REF08a} in the concept discovery stage for its faster running time, and $\chi^2$ kernel SVM to train final concept classifiers as it is a natural choice for histogram-like features and provides higher performance than linear SVM.

\textbf{Comparison against embedding-based approaches:} We first compare the performance of our concept-based pipeline against embedding based approaches.
We set the parameters of our system using the validation set. For concept discovery, we kept all terms with at least 5 occurrences in the training sentences, this gave us an initial list of 5,309 terms. We filtered all terms with average precision lower than 0.15, which preserved 2,877 terms. We set $\lambda$ to be 0.6 and number of concepts to be 1,200.


\begin{table*}
  \begin{center}
  \small
    \begin{tabular}{c|cccc||cccc}
      \hline
      &\multicolumn{4}{c||}{Image to
        sentence}&\multicolumn{4}{c}{Sentence to image}\\
      \hline
      Method&R@1&R@5&R@10&Median rank&R@1&R@5&R@10&Median rank\\
      \hline
      \hline
      Karpathy et al.~\cite{DBLP:journals/corr/KarpathyF14}&16.5&40.6&54.2&\textbf{7.6}&11.8&32.1&44.7&12.4\\
      \hline
      Mao et al.~\cite{DBLP:journals/corr/MaoXYWY14}&14.5&37.2&48.5&11&11.5&31.0&42.4&14\\
      \hline
      Kiros et al.~\cite{DBLP:journals/corr/KirosSZ14}&18.0&40.9&\textbf{55.0}&8&12.5&37.0&51.5&10\\
      \hline
      Concepts (trained on Flickr 8k) & \textbf{18.7}&\textbf{41.9}&54.7&8&\textbf{16.7}&\textbf{40.7}&\textbf{54.0}&\textbf{9}\\
      \hline
      Concepts (trained on Flickr 30k) & \textbf{21.1} & \textbf{45.9} & \textbf{59.0} & \textbf{7} & \textbf{17.9} & \textbf{42.8} & \textbf{55.8} & \textbf{8}\\
      \hline
     \end{tabular}
  \end{center}
  \caption{Retrieval evaluation compared with embedding based methods on Flickr 8k. Higher Recall$@k$ and lower median rank are better.}
  \label{tab:8k_results}
\end{table*}

\begin{table*}
  \begin{center}
  \small
    \begin{tabular}{c|cccc||cccc}
      \hline
      &\multicolumn{4}{c||}{Image to
        sentence}&\multicolumn{4}{c}{Sentence to image}\\
      \hline
      Method&R@1&R@5&R@10&Median rank&R@1&R@5&R@10&Median rank\\
      \hline
      Karpathy et al.~\cite{DBLP:journals/corr/KarpathyF14}&22.2&48.2&61.4&\textbf{4.8}&15.2&37.7&50.5&9.2\\
      \hline
      Mao et al.~\cite{DBLP:journals/corr/MaoXYWY14}&18.4&40.2&50.9&10&12.6&31.2&41.5&16\\
      \hline
      Kiros et al.~\cite{DBLP:journals/corr/KirosSZ14}&23.0&50.7&62.9&5&16.8&42.0&\textbf{56.5}&8\\
      \hline
      Concepts (trained on Flickr 30k) & \textbf{26.6} & \textbf{52.0} & \textbf{63.7} & 5 & \textbf{18.3} & \textbf{42.2} & 56.0 & 8\\
      \hline
     \end{tabular}
  \end{center}
  \caption{Retrieval evaluation on Flickr 30k. Higher Recall$@k$ and lower median rank are better.}
  \label{tab:30k_results}
\end{table*}

Several recent embedding based approaches~\cite{DBLP:journals/corr/KarpathyJF14,DBLP:journals/tacl/SocherKLMN14,DBLP:journals/corr/KarpathyF14,DBLP:journals/corr/MaoXYWY14,DBLP:journals/corr/ChenZ14a,DBLP:journals/corr/KirosSZ14} are included for comparison. Most of these approaches use CNN-based image representations (in particular, \cite{DBLP:journals/corr/KirosSZ14} uses the same Oxford architecture), and embed sentences with recurrent neural network (RNN) or its variations. We make sure that the experiment setup and data partitioning for all systems are the same, and report numbers in the original papers if available.

Table~\ref{tab:8k_results} lists the evaluation performance for all systems. We can see that the concept based framework achieves similar or better performance against the state-of-the-art embedding based systems. This confirms the framework is a valid pipeline for bidirectional image and sentence retrieval task.


\textbf{Enhancing concept classifiers with more data:} The concept classifiers we trained for previous experiment only used training images from Flickr 8k data set. To check if the discovered concepts can benefit from additional training data, we collect the images associated with the discovered concepts from Flickr 30k data set. Since Flickr 30k contains images which overlap with the validation and testing partitions of Flickr 8k data set, we removed those images and used around 29,000 images for training.

In the last row of Table~\ref{tab:8k_results}, we list the results of the concept based approach using Flickr 30k training data. We can see that there is a significant improvement in every metric. Since the only difference is the use of additional training data, the results indicate that \textbf{the individual concept classifiers benefit from extra training data}. It is worth noting that while additional data may also be helpful for embedding based approaches, it has to be in the form of image and sentence pairs. Such annotation tends to be more expensive and time consuming to obtain than concept annotation.

\textbf{Evaluation on Flickr 30k dataset:} Evaluation on Flickr 30k follows the same strategy as on Flickr 8k, where parameters were set using validation data. We kept 9,742 terms which have at least 5 occurrences in the training sentences. We then filtered all terms with average precision lower than 0.15, which preserved 4,158 terms. We set $\lambda$ to be 0.4 and number of concepts to be 1,600. Table~\ref{tab:30k_results} shows that our method achieves comparable or better performance than other embedding based approaches.

\begin{table*}
  \begin{center}
  \small
    \begin{tabular}{c|cccc||cccc}
      \hline
      &\multicolumn{4}{c||}{Image to
        sentence}&\multicolumn{4}{c}{Sentence to image}\\
      \hline
      Vocabulary&R@1&R@5&R@10&Median rank&R@1&R@5&R@10&Median rank\\
      \hline
      ImageNet 1k~\cite{ILSVRCarxiv14}&2.5&6.7&9.7&714&1.6&5.0&8.5&315\\
      \hline
      LEVAN~\cite{DBLP:conf/cvpr/DivvalaFG14}&0.0&0.4&1.2&1348&0.2&1.1&1.7&443\\
      \hline
      NEIL~\cite{chen2013neil}&0.1&0.7&1.1&1103&0.2&0.9&2.0&446\\
      \hline
      LEVAN~\cite{DBLP:conf/cvpr/DivvalaFG14} (trained on Flickr 8k)&1.2&5.7&9.5&360&2.6&9.1&14.7&113\\
      \hline
      NEIL~\cite{chen2013neil} (trained on Flickr 8k)&1.4&5.7&8.9&278&3.7&11.3&18.3&92\\
      \hline
      Flickr 8k Concepts (ours) & \textbf{10.4} & \textbf{29.3} & \textbf{40.0} & \textbf{17} & \textbf{9.8} & \textbf{27.5} & \textbf{39.0} & \textbf{17}\\
      \hline
     \end{tabular}
  \end{center}
  \caption{Retrieval evaluation for different concept vocabularies on COCO data set.}
  \label{tab:coco_results}
\end{table*}

\textbf{Concept transfer to other data sets:} It is important to investigate whether the discovered concepts are generalizable. For this purpose, we randomly selected 1,000 images and their associated 5,000 text descriptions from the validation partition of Microsoft COCO data set~\cite{DBLP:conf/eccv/LinMBHPRDZ14}.

We used the concepts discovered and trained from Flickr 8k data set, and compared with several existing concept vocabularies:

\textbf{ImageNet 1k}~\cite{ILSVRCarxiv14} is a subset of ImageNet data set, with 1,000 categories used in ILSVRC 2014 evaluation. The classifiers were trained using the same Oxford CNN architecture used for feature extraction.

\textbf{LEVAN}~\cite{DBLP:conf/cvpr/DivvalaFG14} selected 305 concepts manually, and explored Google Ngram data to collect 113,983 sub-concepts. They collected Internet images and trained detectors with Deformable Part Model (DPM). We used the learned models provided by the authors.

\textbf{NEIL}~\cite{chen2013neil} has 2,702 manually selected concepts, each of which was trained with DPM using weakly supervised images from search engines. We also used the models released by the authors.

Among the three baselines above, ImageNet 1k relies on the same set of CNN-based features as our discovered concepts. To further investigate the effect of concept selection, we took the concept lists provided by the authors of LEVAN and NEIL, and re-trained their concept detectors using our proposed pipeline. To achieve this, we selected training images associated with the concepts from Flickr 8k dataset, and learned concept detectors using the same CNN feature extractors and classifier training strategies as our proposed pipeline.

Table~\ref{tab:coco_results} lists the performance of using different vocabularies. We can see that \textbf{the discovered concepts clearly outperform manually selected vocabularies}, but \textbf{the cross-dataset performance is lower than same-dataset performance}. We found that COCO uses many visual concepts discovered in Flickr 8k, though some are missing (e.g. \textit{giraffes}). Compared with the concepts discovered by Flickr 8k, the three manually selected vocabularies lack many terms used in the COCO data set to describe the visual entities. This inevitably hurts their performance in the retrieval task. The performance of NEIL and LEVAN is worse than ImageNet 1k, which might be explained by the weakly Internet images they used to train concept detectors. Although re-training from Flickr 8k using deep features helps improve retrieval performance of NEIL and LEVAN, our system still outperforms the two by large margins.

\begin{figure}
\begin{center}
\begin{tabular}{c}
\includegraphics[width=0.9\linewidth]{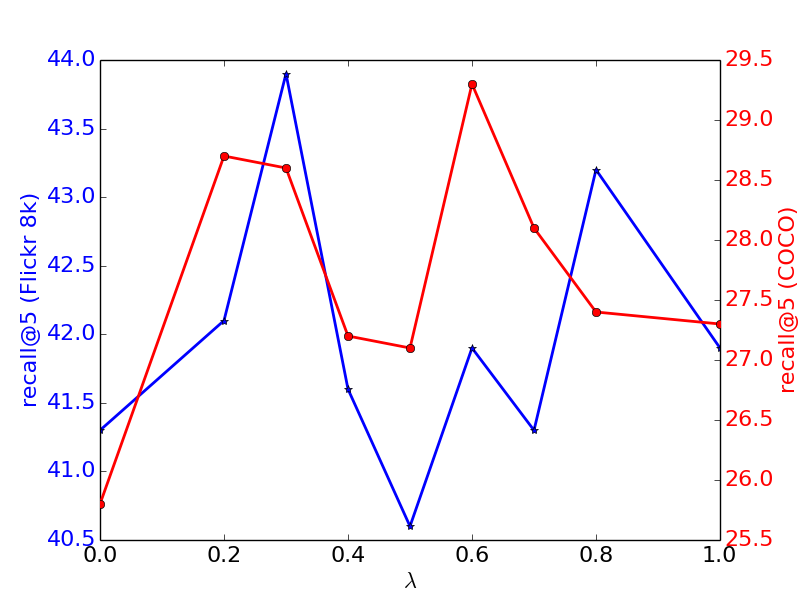}\\
\end{tabular}
\end{center}
\caption{Impact of $\lambda$ when testing on Flickr 8k data set (blue) and COCO data set (red). Recall$@5$ for sentence retrieval is used.}
\label{fig:lambda}
\end{figure}

\begin{figure}
\begin{center}
\begin{tabular}{c}
\includegraphics[width=0.9\linewidth]{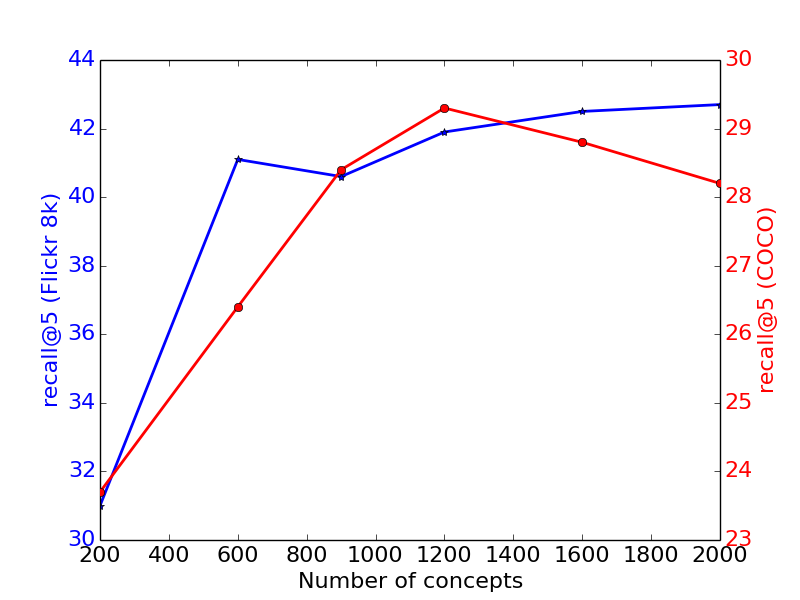}\\
\end{tabular}
\end{center}
\caption{Impact of total number of concepts when testing on Flickr 8k data set (blue) and COCO data set (red). Recall$@5$ for sentence retrieval is used.}
\label{fig:num_cluster}
\end{figure}

\textbf{Impact of concept discovery parameters:} Figure~\ref{fig:lambda} and Figure~\ref{fig:num_cluster} shows the impact of visual similarity weight $\lambda$ and the total number of concepts on the retrieval performance. To save space, we only display results of recall$@5$ for the sentence retrieval direction.

We can see from the figures that both visual and semantic similarities are important for concept clustering, this is particular true when the concepts trained from Flickr 8k were applied to COCO. Increasing the number of concepts helps at the beginning, as many visually discriminative concepts are grouped together when the number of concepts is small. However, as the number increases, the improvement becomes flat, and even hurts the concepts' ability to generalize.

\subsection{Human Evaluation of Image Tagging}
We also evaluated the quality of the discovered concepts on the image tagging task whose goal is to generate tags to describe the content of images. Compared with sentence retrieval, the image tagging task has a higher degree of freedom as the combination of tags is not limited by the existing sentences in the pool.

\textbf{Evaluation setup:} We used the concept classifiers to generate image tags. For each image, we selected the top three concepts with highest classifier scores. Since a concept may have more than one text terms, we picked up to two text terms randomly for display.

For evaluation, we asked 15 human evaluators to compare two sets of tags generated by different concept vocabularies. The evaluators were asked to select which set of tags better describes the image based on the accuracy of the generated tags and the coverage of visual entities in the image, or whether the two sets of tags are equally good or bad. The final label per image was combined using majority vote. On average, 85\% of the evaluators agreed on their votes for specific images.

We compared the concepts discovered from Flickr 8k and the manually selected ImageNet 1k concept vocabulary. The classifiers for the discovered concepts were trained using the 6,000 images from Flickr 8k. We did not compare the discovered concepts against NEIL and LEVAN as they performed very poorly in the retrieval task. To test how the concepts generalize to a different data set, we used the same 1,000 images from the COCO data set as used in retrieval task for evaluation.

\begin{table}
  \begin{center}
    \begin{tabular}{ccc}
      \hline
      Better & Worse & Same\\
      \hline
      \textbf{64.1}\% & 22.9\% & 12.9\%\\
      \hline
     \end{tabular}
  \end{center}
  \caption{Percentage of images where tags generated by the discovered concepts are better, worse or the same compared with ImageNet 1k.}
  \label{tab:human_eval}
\end{table}

\begin{figure}
\begin{center}
\begin{tabular}{c}
\includegraphics[width=0.95\linewidth]{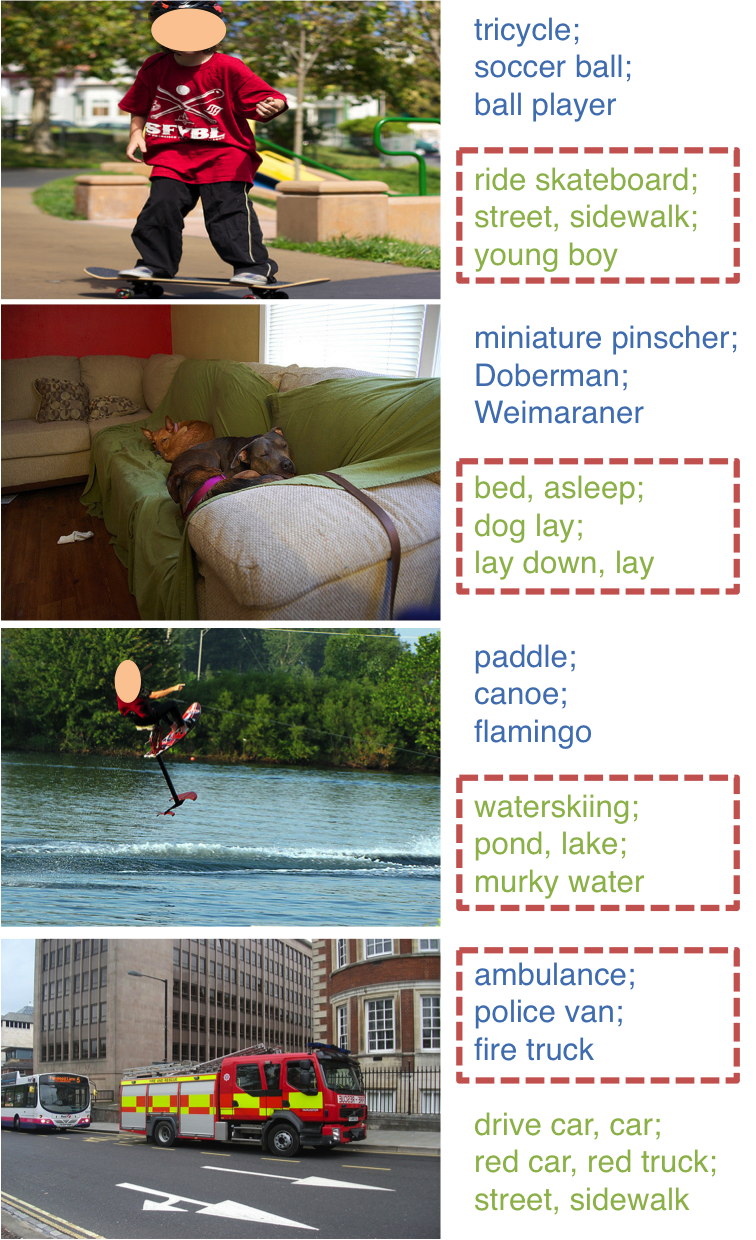}\\
\end{tabular}
\end{center}
\caption{Tags generated using ImageNet 1k concepts (blue) and the discovered concepts (green). Tags preferred by evaluators are marked in red blocks.}
\label{fig:human_eval}
\end{figure}

\textbf{Result analysis:} Table~\ref{tab:human_eval} shows the evaluators' preference on the image tags generated by the discovered concepts and ImageNet 1k. We can see that the discovered concepts generated better tags for 64.1\% of the images. This agrees with the trend observed in the bidirectional retrieval task.


As shown in Figure~\ref{fig:human_eval}, tags generated by ImageNet 1k has the following problems which might cause evaluators to label them as worse: first, 
many of the visual entities do not have corresponding concepts in the vocabulary; second, ImageNet 1k has many fine-grained concepts (e.g. different species of dogs), while more general terms might be preferred by evaluators. 
On the other hand, the discovered concepts are able to reflect how human name the visual entities, and have a higher concept coverage. However, due to the number of training examples is relatively limited, sometimes the response of different concept classifiers are correlated (e.g. \textit{bed} and \textit{sit down}).

\section{Conclusion}
\label{sec:conclusion}
This paper studies the problem of automatic concept discovery from
parallel corpora. We propose a concept filtering and clustering algorithm using both text and visual information. Automatic evaluation using bidirectional image and text retrieval and human evaluation of image tagging task show that the discovered concepts achieve state-of-the-art performance, and outperform several large manually selected concept vocabularies significantly. A natural future direction is to train concept detectors for the discovered concepts using web images.

{
\quad\newline
\textbf{Acknowledgement:}
We thank Kevin Knight for helpful discussions.
This work was supported by the Intelligence Advanced Research Projects Activity (IARPA) via Department of Interior National Business Center contract number D11PC20066. The U.S. Government is authorized to reproduce and distribute reprints for Governmental purposes notwithstanding any copyright annotation thereon.
Disclaimer: The views and conclusions contained herein are those of the authors and should not be interpreted as necessarily representing the official policies or endorsements, either expressed or implied, of IARPA, DoI/NBC, or the U.S. Government.
}

{\small
\bibliographystyle{ieee}
\bibliography{egbib}

\begin{thebibliography}{10}\itemsep=-1pt

\bibitem{DBLP:dblp_journals/jmlr/BachJ02}
F.~R. Bach and M.~I. Jordan.
\newblock Kernel independent component analysis.
\newblock {\em JMLR}, 2002.

\bibitem{chen2013watching}
C.-Y. Chen and K.~Grauman.
\newblock Watching unlabeled video helps learn new human actions from very few
  labeled snapshots.
\newblock In {\em CVPR}, 2013.

\bibitem{chen2013neil}
X.~Chen, A.~Shrivastava, and A.~Gupta.
\newblock {NEIL}: Extracting visual knowledge from web data.
\newblock In {\em ICCV}, 2013.

\bibitem{DBLP:journals/corr/ChenZ14a}
X.~Chen and C.~L. Zitnick.
\newblock Learning a recurrent visual representation for image caption
  generation.
\newblock {\em CoRR}, abs/1411.5654, 2014.

\bibitem{Marneffe06generatingtyped}
M.-C. de~Marneffe, B.~MacCartney, and C.~D. Manning.
\newblock Generating typed dependency parses from phrase structure parses.
\newblock In {\em LREC}, 2006.

\bibitem{imagenet_cvpr09}
J.~Deng, W.~Dong, R.~Socher, L.-J. Li, K.~Li, and L.~Fei-Fei.
\newblock {ImageNet: A Large-Scale Hierarchical Image Database}.
\newblock In {\em CVPR}, 2009.

\bibitem{DengKrauseBergFei-Fei_CVPR2012}
J.~Deng, J.~Krause, A.~Berg, and L.~Fei-Fei.
\newblock Hedging your bets: Optimizing accuracy-specificity trade-offs in
  large scale visual recognition.
\newblock In {\em CVPR}, 2012.

\bibitem{DBLP:conf/cvpr/DivvalaFG14}
S.~K. Divvala, A.~Farhadi, and C.~Guestrin.
\newblock Learning everything about anything: Webly-supervised visual concept
  learning.
\newblock In {\em CVPR}, 2014.

\bibitem{caba2015activitynet}
B.~G. Fabian Caba~Heilbron, Victor~Escorcia and J.~C. Niebles.
\newblock {ActivityNet}: A large-scale video benchmark for human activity
  understanding.
\newblock In {\em CVPR}, 2015.

\bibitem{REF08a}
R.-E. Fan, K.-W. Chang, C.-J. Hsieh, X.-R. Wang, and C.-J. Lin.
\newblock {LIBLINEAR}: A library for large linear classification.
\newblock {\em JMLR}, 2008.

\bibitem{DBLP:conf/cvpr/FarhadiEHF09}
A.~Farhadi, I.~Endres, D.~Hoiem, and D.~A. Forsyth.
\newblock Describing objects by their attributes.
\newblock In {\em CVPR}, 2009.

\bibitem{devise}
A.~Frome, G.~Corrado, J.~Shlens, S.~Bengio, J.~Dean, M.~Ranzato, and
  T.~Mikolov.
\newblock Devise: A deep visual-semantic embedding model.
\newblock In {\em NIPS}, 2013.

\bibitem{girshick14CVPR}
R.~Girshick, J.~Donahue, T.~Darrell, and J.~Malik.
\newblock Rich feature hierarchies for accurate object detection and semantic
  segmentation.
\newblock In {\em CVPR}, 2014.

\bibitem{DBLP:conf/eccv/GolgeD14}
E.~Golge and P.~Duygulu.
\newblock Conceptmap: Mining noisy web data for concept learning.
\newblock In {\em ECCV}, 2014.

\bibitem{DBLP:conf/eccv/GongWHHL14}
Y.~Gong, L.~Wang, M.~Hodosh, J.~Hockenmaier, and S.~Lazebnik.
\newblock Improving image-sentence embeddings using large weakly annotated
  photo collections.
\newblock In {\em ECCV}, 2014.

\bibitem{guadarrama:ICCV13}
S.~Guadarrama, N.~Krishnamoorthy, G.~Malkarnenkar, R.~Mooney, T.~Darrell, and
  K.~Saenko.
\newblock {YouTube2Text}: Recognizing and describing arbitrary activities using
  semantic hierarchies and zero-shot recognition.
\newblock In {\em ICCV}, 2013.

\bibitem{DBLP:journals/jair/HodoshYH13}
M.~Hodosh, P.~Young, and J.~Hockenmaier.
\newblock Framing image description as a ranking task: Data, models and
  evaluation metrics.
\newblock {\em JAIR}, 2013.

\bibitem{jia2014caffe}
Y.~Jia, E.~Shelhamer, J.~Donahue, S.~Karayev, J.~Long, R.~Girshick,
  S.~Guadarrama, and T.~Darrell.
\newblock Caffe: Convolutional architecture for fast feature embedding.
\newblock In {\em ACM MM}, 2014.

\bibitem{DBLP:journals/corr/KarpathyF14}
A.~Karpathy and L.~Fei{-}Fei.
\newblock Deep visual-semantic alignments for generating image descriptions.
\newblock {\em CVPR}, 2015.

\bibitem{DBLP:journals/corr/KarpathyJF14}
A.~Karpathy, A.~Joulin, and L.~Fei{-}Fei.
\newblock Deep fragment embeddings for bidirectional image sentence mapping.
\newblock In {\em NIPS}, 2014.

\bibitem{DBLP:journals/corr/KirosSZ14}
R.~Kiros, R.~Salakhutdinov, and R.~S. Zemel.
\newblock Unifying visual-semantic embeddings with multimodal neural language
  models.
\newblock {\em TACL}, 2015.

\bibitem{AlexNet}
A.~Krizhevsky, I.~Sutskever, and G.~E. Hinton.
\newblock Imagenet classification with deep convolutional neural networks.
\newblock In {\em NIPS}, 2012.

\bibitem{Kulkarni11babytalk}
G.~Kulkarni, V.~Premraj, S.~Dhar, S.~Li, Y.~Choi, A.~C. Berg, and T.~L. Berg.
\newblock Baby talk: Understanding and generating image descriptions.
\newblock In {\em CVPR}, 2011.

\bibitem{kuznetsova2012collective}
P.~Kuznetsova, V.~Ordonez, A.~C. Berg, T.~L. Berg, and Y.~Choi.
\newblock Collective generation of natural image descriptions.
\newblock In {\em ACL}, 2012.

\bibitem{DBLP:conf/acl/KuznetsovaOBBC13}
P.~Kuznetsova, V.~Ordonez, A.~C. Berg, T.~L. Berg, and Y.~Choi.
\newblock Generalizing image captions for image-text parallel corpus.
\newblock In {\em ACL}, 2013.

\bibitem{LiSXL10OB}
L.-J. Li, H.~Su, E.~P. Xing, and F.-F. Li.
\newblock Object bank: A high-level image representation for scene
  classification {\&} semantic feature sparsification.
\newblock In {\em NIPS}, 2010.

\bibitem{DBLP:conf/eccv/LinMBHPRDZ14}
T.~Lin, M.~Maire, S.~Belongie, J.~Hays, P.~Perona, D.~Ramanan, P.~Doll{\'{a}}r,
  and C.~L. Zitnick.
\newblock Microsoft {COCO:} common objects in context.
\newblock In {\em ECCV}, 2014.

\bibitem{DBLP:journals/corr/MaoXYWY14}
J.~Mao, W.~Xu, Y.~Yang, J.~Wang, and A.~L. Yuille.
\newblock Explain images with multimodal recurrent neural networks.
\newblock {\em CoRR}, abs/1410.1090, 2014.

\bibitem{word2vec}
T.~Mikolov, I.~Sutskever, K.~Chen, G.~S. Corrado, and J.~Dean.
\newblock Distributed representations of words and phrases and their
  compositionality.
\newblock In {\em NIPS}, 2013.

\bibitem{Miller95wordnet:a}
G.~A. Miller.
\newblock {WordNet: A Lexical Database for English}.
\newblock {\em CACM}, 1995.

\bibitem{DBLP:conf/iccv/OrdonezDCBB13}
V.~Ordonez, J.~Deng, Y.~Choi, A.~C. Berg, and T.~L. Berg.
\newblock From large scale image categorization to entry-level categories.
\newblock In {\em ICCV}, 2013.

\bibitem{ordonez2011im2text}
V.~Ordonez, G.~Kulkarni, and T.~L. Berg.
\newblock Im2text: Describing images using 1 million captioned photographs.
\newblock In {\em NIPS}, 2011.

\bibitem{rohrbachtranslating}
M.~Rohrbach, W.~Qiu, I.~Titov, S.~Thater, M.~Pinkal, and B.~Schiele.
\newblock Translating video content to natural language descriptions.
\newblock In {\em ICCV}, 2013.

\bibitem{rosch1978}
E.~Rosch.
\newblock Principles of categorization.
\newblock 1978.

\bibitem{ILSVRCarxiv14}
O.~Russakovsky, J.~Deng, H.~Su, J.~Krause, S.~Satheesh, S.~Ma, Z.~Huang,
  A.~Karpathy, A.~Khosla, M.~Bernstein, A.~C. Berg, and L.~Fei-Fei.
\newblock {ImageNet Large Scale Visual Recognition Challenge}, 2014.

\bibitem{Sadanand12AB}
S.~Sadanand and J.~Corso.
\newblock Action bank: A high-level representation of activity in video.
\newblock In {\em CVPR}, 2012.

\bibitem{DBLP:conf/cvpr/SadeghiF11}
M.~A. Sadeghi and A.~Farhadi.
\newblock Recognition using visual phrases.
\newblock In {\em CVPR}, 2011.

\bibitem{DBLP:journals/corr/SimonyanZ14a}
K.~Simonyan and A.~Zisserman.
\newblock Very deep convolutional networks for large-scale image recognition.
\newblock {\em NIPS}, 2014.

\bibitem{DBLP:journals/tacl/SocherKLMN14}
R.~Socher, A.~Karpathy, Q.~V. Le, C.~D. Manning, and A.~Y. Ng.
\newblock Grounded compositional semantics for finding and describing images
  with sentences.
\newblock {\em {TACL}}, 2014.

\bibitem{sun_nevatia_cvpr14}
C.~Sun and R.~Nevatia.
\newblock {DISCOVER}: Discovering important segments for classification of
  video events and recounting.
\newblock In {\em CVPR}, 2014.

\bibitem{SunECCV}
C.~Sun and R.~Nevatia.
\newblock Semantic aware video transcription using random forest classifiers.
\newblock In {\em ECCV}, 2014.

\bibitem{wu2015webconcept}
J.~Wu, Y.~Yu, C.~Huang, and K.~Yu.
\newblock Deep multiple instance learning for image classification and
  auto-annotation.
\newblock {\em CVPR}, 2015.

\bibitem{DBLP:journals/tacl/YoungLHH14}
P.~Young, A.~Lai, M.~Hodosh, and J.~Hockenmaier.
\newblock From image descriptions to visual denotations: New similarity metrics
  for semantic inference over event descriptions.
\newblock {\em {TACL}}, 2014.

\bibitem{zhou2015conceptlearner}
B.~Zhou, V.~Jagadeesh, and R.~Piramuthu.
\newblock {ConceptLearner: Discovering Visual Concepts from Weakly Labeled
  Image Collections.}
\newblock {\em CVPR}, 2015.

\bibitem{zhou2014places}
B.~Zhou, A.~Lapedriza, J.~Xiao, A.~Torralba, and A.~Oliva.
\newblock {Learning Deep Features for Scene Recognition using Places Database.}
\newblock {\em NIPS}, 2014.

\end{thebibliography}
}

\end{document}